\pdfoutput=1
\documentclass[11pt]{article}

\usepackage[final]{acl}

\usepackage{times}
\usepackage{latexsym}

\usepackage[T1]{fontenc}
\usepackage[utf8]{inputenc}

\usepackage{microtype}

\usepackage{inconsolata}

\usepackage{graphicx}

\usepackage{amsmath}
\usepackage{amssymb}
\usepackage{multirow}
\usepackage{arydshln}
\usepackage{graphicx}
\usepackage{wrapfig}
\usepackage{colortbl}
\usepackage{booktabs}
\usepackage{makecell}
\newcommand{\bftab}{\fontseries{b}\selectfont}
\usepackage{diagbox}
\usepackage{float}

\RequirePackage{xspace}
\makeatletter
\DeclareRobustCommand\onedot{\futurelet\@let@token\@onedot}
\def\@onedot{\ifx\@let@token.\else.\null\fi\xspace}

\def\eg{\emph{e.g}\onedot}
\def\ie{\emph{i.e}\onedot}
\title{ModalPrompt: Towards Efficient Multimodal Continual Instruction Tuning with Dual-Modality Guided Prompt}

\author{
 \textbf{Fanhu Zeng\textsuperscript{\rm 1,2}},
 \textbf{Fei Zhu\textsuperscript{\rm 3}},
 \textbf{Haiyang Guo\textsuperscript{\rm 1}},
 \textbf{Xu-Yao Zhang\textsuperscript{\rm 1,2}\thanks{Corresponding Author.}},
 \textbf{Cheng-Lin Liu\textsuperscript{\rm 1,2}} \\
 {\normalsize \textsuperscript{1}State Key Laboratory of Multimodal Artificial Intelligence Systems, CASIA} \\
 {\normalsize \textsuperscript{2}School of Artificial Intelligence, UCAS} 
 {\normalsize \textsuperscript{3}Centre for Artificial Intelligence and Robotics, HKISI-CAS } \\
 {\normalsize \{zengfanhu2022, guohaiyang2023\}@ia.ac.cn, \ zhfei2018@gmail.com, \ \{xyz, liucl\}@nlpr.ia.ac.cn}
}

\begin{document}
\maketitle
\begin{abstract}
Large Multimodal Models~(LMMs) exhibit remarkable multi-tasking ability by learning mixed instruction datasets. However, novel tasks would be encountered sequentially in dynamic world, which urges for equipping LMMs with multimodal continual instruction learning~(MCIT) ability especially for diverse and challenging generative tasks. Existing MCIT methods do not fully exploit the unique attribute of LMMs and often gain performance at the expense of efficiency. In this paper, we propose a novel prompt learning framework for MCIT to effectively alleviate forgetting of previous knowledge while managing computational complexity with natural image-text supervision. Concretely, we learn prompts for each task and exploit efficient prompt fusion for knowledge transfer and prompt selection for complexity management with dual-modality guidance. Extensive experiments demonstrate that our approach achieves substantial \textbf{+14.26\%} performance gain on MCIT benchmarks with remarkable $\times$\textbf{1.42} inference speed free from growing computation. Code is available at~\url{https://github.com/AuroraZengfh/ModalPrompt}.
\end{abstract}

\section{Introduction}
\label{sec:intro}
In recent years, large multimodal model~(LMM), which aligns visual encoder~\cite{dosovitskiy2020image} with large language model~(LLM) to handle multimodal tasks, has gained remarkable performance in numerous fields~\cite{li2023blip2,liu2024visual}. As models become larger~\cite{dubey2024llama, zeng2025parameter, zeng2025local}, they are expected to perform lifelong learning like humans and learn more than one time to handle multiple tasks other than single tasks~\cite{yao2022filip, dai2024instructblip}.

However, while pre-trained models like LLaVA perform well on mixed datasets, they tend to forget older tasks when fine-tuned on new task. Such catastrophic forgetting phenomenon is especially evident in sequential learning of widely differing multimodal tasks such as VQA and grounding~\cite{goyal2017making,deng2021transvg}. This calls for multimodal continual instruction tuning~(MCIT), which aims at sequentially fine-tuning models with multimodal instruction datasets and gets superior performance on new tasks while maintaining ability on previous tasks.

\begin{figure}[t]
    \centering
    \includegraphics[width=\linewidth]{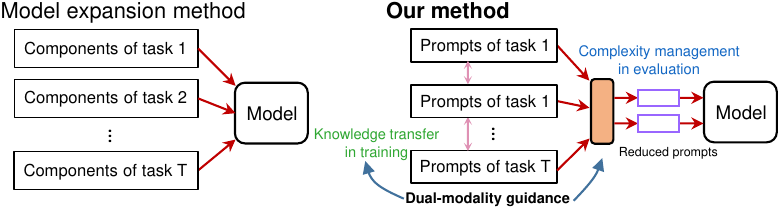}
    \vspace{-15pt}
    \caption{Diagram of model expansion method and our method. With natural attribute of multimodal guidance, we enhance MCIT with knowledge transfer and manage complexity against linear growth.}
    \label{fig:comparison}
    \vspace{-18pt}
\end{figure}

Existing approaches mainly tackle the forgetting issue by continually extending model with separate lightweight component for each task shown in Fig.~\ref{fig:comparison}, and LoRA~\cite{hu2022lora} appears to be the common practice for large models~\cite{wang2023orthogonal}. However, they expand model size and inference time in proportion to the number of tasks since they ensemble separated components of each task during inference. As the number of tasks increases, the cost of storage and inference become unbearable, particularly in LMMs and therefore hinder their practical deployments in real-world scenarios. Moreover, current methods derived from language models are not specially designed for LMMs~\cite{hu2023pop, razdaibiedina2023progressive} without fully exploiting information from vision side and inevitably perform poorly on multimodal benchmarks. The mentioned shortcomings naturally raise an open question: \emph{Can we establish an effective MCIT framework designed for LMMs refraining from growing computational expansion?}

In this paper, we investigate how to retain information of older tasks from multimodality~(\ie, image and text) to fully exploit LMMs and therefore improve the performance of multimodal continual instruction tuning efficiently. Generally speaking, given that the primary distinction between LLM and LMM lies in their utilization of image features, we establish a general prompt learning framework for multimodal continual instruction tuning with supervision from multimodality. \textbf{First}, we build a set of prompts for each task to represent task-specific knowledge and a lightweight projection layer is exploited to extract \textit{prototype features} from task-specific prompts. When multimodal inputs come, the off-the-shelf text and visual encoders are used to obtain \textit{multimodal features} representing multimodal distribution of current input. \textit{Prototype} and \textit{multimodal features} are then matched with similarity denoted as \textit{dual-modality guidance}. \textbf{Second}, to enhance knowledge transfer, prompts that are most relevant to current task are obtained and fused through dual-modality guidance to promote the performance. \textbf{Third}, to address the problem of computational complexity, prompt selection mechanism from dual-modality guidance is developed to maintain inference efficiency.

Our method has two advantages: \textbf{(1)} guidance features after tokenization~(text) and projection~(image) naturally align multimodality information and are effortless to guide knowledge transfer and selection of LMMs; \textbf{(2)} computational complexity is in proportion to token numbers other than task numbers, and can therefore manage time consumption by selecting proper tokens. Extensive evaluation on MCIT benchmark across diverse multimodal tasks validates that our method substantially boosts performance on older tasks and mitigates forgetting with great training and inference efficiency. Our contributions are summarized as follows:

\begin{itemize}
    \item We propose ModalPrompt, the first prompt learning framework for multimodal continual instruction tuning to mitigate forgetting with the advantage of multimodal supervision.
    \item We construct prompts to retain knowledge of specific tasks and exploit an effective dual-modality guided prompt fusion and selection technique to ensure MCIT ability while managing computational complexity.
    \item We conduct extensive experiments on continual instruction tuning benchmark, and the results substantially outperform existing methods~(\textbf{+14.26\%}) with great efficiency.
\end{itemize}

\section{Related Work}
\label{sec:related}
\textbf{Large Multimodal Models}~(LMMs)~\cite{liu2024visual,liu2024improved,ye2024mplug}, which combine vision representation with large language models~(LLMs)~\cite{alayrac2022flamingo,touvron2023llama}, have exhibited predominant function in numerous multimodal tasks~\cite{liu2023mmbench,lu2024mathvista}. They typically contain a LLM decoder with stacks of transformers to decode embeddings. Usually, they first process image pixels with a CLIP image encoder, align features with a linear projector and then generate responses with concatenation of both image-text representations in an autoregressive way as LLMs do. As full fine-tuning is time-consuming and resource-intensive, efficient tuning is the common practice for instruction tuning of large models~\cite{han2024parameter}. Methods for parameter efficient tuning are mainly three-fold: adapter learning~\cite{zhang2021tip,satapara2024tl,lee2024incremental}, prompt learning~\cite{zhou2022learning} and LoRA~\cite{hu2022lora}. They update models with a lightweight module in the form of intra-block parallel connections, prefixes among input embeddings and low-rank decomposition, respectively. Specifically, multimodal instruction tuning~\cite{wang2024m2pt, liu2025re} has been a promising direction in promoting performance of multimodal models with both LoRA~\cite{shen2024multimodal, xu2024lateralization} and prompt learning~\cite{khattak2023maple}. As an orthogonal direction, we exploit techniques for mitigating catastrophic forgetting in multimodal foundation models.

\begin{figure*}[ht]
    \centering
    \includegraphics[width=0.93\linewidth]{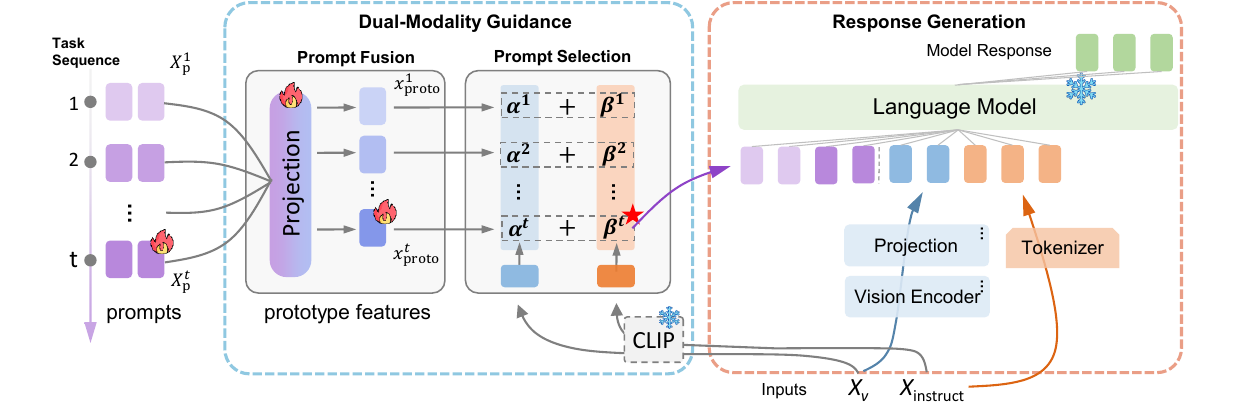}
    \vspace{-10pt}
    \caption{\emph{Left:} prompt fusion module. Prototype features ($x_{\mathrm{proto}}^t$) are obtained from the projection of prompts ($X_\mathrm{p}^t$) to get task-specific knowledge in feature space. \emph{Middle:} prompt selection process. Prototype features that are the most similar to current multimodal features are selected to enhance current input. \emph{Right:} selected prompts and original multimodal inputs are concatenated and fed into large language model to generate responses.}
    \vspace{-15pt}
    \label{fig:detailed-structure}
\end{figure*}

\noindent \textbf{Continual Instruction Tuning}~\cite{guo2025comprehensive} goes beyond instruction tuning that adapts large models to understand and align with human instructions. It primarily solves the problem of catastrophic forgetting~\cite{zhai2023investigating,guo2025hide,guo2025pilora} when one large foundation model sequentially learns multiple tasks through instruction tuning datasets. With the extensive development of LMM, much attention and effort have been paid to multimodal continual instruction tuning~\cite{wu2024continual,zhang2024mm}. However, mainstream methods focus on transferring CIT methods from language tasks~\cite{wang2024inscl,wang2024rehearsal,wang2023orthogonal} with no special design for visual features~\cite{he2023continual}. Recently, a multimodal continual instruction tuning benchmark named CoIN~\cite{chen2024coin} has been established and MoELoRA~\cite{dou2023loramoe} is adopted to align previous instructions. However, it suffers from severe performance drop, highlighting the necessity for exploration solutions tailored for multimodal continual instruction tuning.

\section{Method: ModalPrompt}
\label{sec:preliminary}
\noindent \textbf{Notations.} Multimodal continual instruction tuning seeks to address the issue of empowering LMM with ongoing datasets, where LMM $f_\theta(\cdot)$ is pre-trained on large-scale vision-language data to align image-text features. Given $T$ tasks $\{\mathcal{T}^1,\cdots,\mathcal{T}^T\}$ with corresponding multimodal data $\mathcal{D}_t =\{X_\mathrm{v}^{t,i},X_\mathrm{instruct}^{t,i},y^{t,i} \}_{i=1}^{N_t} , t=1,\cdots,T$, where $X_\mathrm{v}^{t,i}$, $X_\mathrm{instruct}^{t,i}$, $y^{t,i}$ stand for $i^{th}$ sample of image, text and ground truth for $t^{th}$ dataset~($N_t$ in total), respectively. A continual learner aims to fine-tune $f_\theta(\cdot)$ sequentially on current data $D_t$ while retaining knowledge on all previous tasks $\mathcal{T}^{< t}$~\footnote{we use the superscript for all elements from 1 to $t$.}. For clarification, Tab.~\ref{tab:notations} summarizes notations that would be used widely in this paper.

\begin{table}[t]
    \centering
    \vspace{4pt}
    \resizebox{\linewidth}{!}{
    \begin{tabular}{cc}
    \toprule[1.3pt]
    \textbf{Notation} & \textbf{Explanation} \\\midrule
        $X_\mathrm{p}^t$  & Prompts for task $t$ \\\midrule
        $x_\mathrm{proto}^t$ & Prototype features of prompts for task $t$  \\\midrule
        $X_\mathrm{v}$, $X_\mathrm{insturct}$ & Image and text inputs of instruction tuning\\\midrule
        $x_\mathrm{v}$, $x_\mathrm{instruct}$ & Image and text features of guidance\\\midrule
        $\alpha^t$, $\beta^t$ & Guidance from image and text modalities\\\midrule
        $\widetilde{X}_\mathrm{p}^t$, $\widetilde{X}_\mathrm{p}^{\mathrm{eval}}$ & Selected prompts for training task $t$ and evaluation\\
    \bottomrule[1.3pt]
    \end{tabular}}
    \vspace{-7pt}
    \caption{Explanations of notations.}
    \vspace{-17pt}
    \label{tab:notations}
\end{table}

\vspace{0.2em}
\noindent \textbf{Problem setup.} 
In this paper, we focus on multimodal continual instruction tuning in a more practical and challenging setting: (1) \textbf{Diverse generative tasks}: continual learning procedure is focused on generative tasks other than simple discriminative tasks like classification and with existence of vision information, task type is much more diverse with abundant scenarios; (2) \textbf{Free from task identification}: during inference, the model does not possess prior knowledge regarding which specific task current question belongs to; (3) \textbf{Absence of replay samples}: due to data privacy, no raw samples are replayed to refresh knowledge of previous tasks.

\vspace{0.2em}
\noindent \textbf{Overview.} 
We present the basic prompt learning framework for multimodal continual instruction tuning. As illustrated in right of Fig.~\ref{fig:detailed-structure}, the structure is similar to normal LMM, other than the input is prefixed with several prompts representing task-specific knowledge presented on the left. Given a set of prompts $X_\mathrm{p}^t$ with length $M$ attached to each task $t, t\in \{1,\cdots,T\}$ representing task-specific knowledge in the form of MCIT, we focus on \textbf{(1) knowledge transfer} and \textbf{(2) complexity management}, which are \textbf{connected by dual-modality guidance}. Specifically, with dual-modality guidance matching multimodal input and task-specific knowledge, we propose multi-task prompt fusion to enhance knowledge transfer among different tasks during training in Sec.~\ref{sec:prompt-fusion} and focus on how to manage complexity by prompt selection at inference time in Sec.~\ref{sec:prompt-selection}.

\subsection{Dual-Modality Feature as Guidance}
The core for MCIT is how to bring \textbf{multimodal knowledge} of similar tasks to current input and generate response reasonably. Given that multimodal input naturally brings image-text supervision, we aim to match input and prompt in multimodal space. For image $X_\mathrm{v}$ and text $X_\mathrm{instruct}$ in each input of current task, considering that CLIP well captures image-text features, we reuse off-the-shelf vision and text encoder from CLIP to extract \textit{multimodal features} of specific input:
\begin{equation}
    \small
    x_\mathrm{v} = \mathrm{Proj_v}(E_I(X_\mathrm{v})), \quad  x_\mathrm{instruct} = E_T(X_\mathrm{instruct}),
    \label{eqn:dual-guidance}
\end{equation}
where $E_I(\cdot): \mathbb{R}^{\mathrm{n_v} \times d_\mathrm{v}} \rightarrow \mathbb{R}^{d_\mathrm{v}}, E_T(\cdot):\mathbb{R}^{\mathrm{n_t}\times {d_\mathrm{t}}} \rightarrow \mathbb{R}^{d_\mathrm{t}}$ and $\mathrm{Proj_v(\cdot)}:\mathbb{R}^{d_\mathrm{v}} \rightarrow \mathbb{R}^{d_\mathrm{t}}$ are CLIP vision encoder, text encoder and linear projection, respectively. $\mathrm{n_v}, \mathrm{n_t}, \mathrm{d_v}$ and $\mathrm{d_t}$ are length of image and text inputs, visual and textual dimension. The utilization can be effortless as they are well-trained and frozen for feature extraction, and the vision encoder has already been used in LMM to extract image features. We give detailed analysis of different encoders in Appendix~\ref{app:full-results}. The extracted text and visual features is crucial in enhancing continual ability~(Sec.~\ref{sec:prompt-fusion}) and managing complexity~(Sec.~\ref{sec:prompt-selection}) described subsequently.

\subsection{Training: Multi-Task Prompt Fusion}
\label{sec:prompt-fusion}
In contrast to class incremental learning~\cite{zhu2021prototype, guo2024desire, wang2022learning} that learns distinct categories, datasets for MCIT like multiple types of question-answering tasks share general knowledge. Without explicit knowledge sharing among prompts of different tasks, in what follows, we first propose to transfer similar knowledge of older tasks in training procedure through multi-task prompt fusion to explicitly enhance MCIT. As illustrated in left of Fig.~\ref{fig:detailed-structure}, prompts of all previous tasks are frozen for knowledge reuse and only current prompts are trainable, and we continually integrate knowledge of older tasks during sequential instruction tuning of current task with the aid of dual-modality features.

\noindent \textbf{Prompt fusion for knowledge transfer.} When training the $t^{th}$ task, the trainable prompts are supposed to draw close to vision-language features of current task and absorb potential knowledge that may boost the performance. To enhance knowledge transfer, the dual-modality features could serve as guiding cues for prompts to accurately get close to multimodal distributions of current task in feature space. We propose to build prototype features from a lightweight projection layer~(\eg, MLP) to further align task-specific knowledge with guidance features from input:
\begin{equation}
   x_\mathrm{proto}^t =  \mathrm{Proj_p}(X_\mathrm{p}^t),
   \label{eqn:proto}
\end{equation}
where $\mathrm{Proj_p}(\cdot)$: $\mathbb{R}^{M\times d_t} \rightarrow \mathbb{R}^{d_t}$ projects the prompts into task-specific \textit{prototype features} in image-text feature space. It is effective in distinguishing whether prompts of older tasks are favorable for current tasks. Then, we explicitly match prompts and current input by fusing the prompts with the largest similarity of multimodal supervision for knowledge transfer. Concretely, we exploit the similarity between prototype features and dual-modality features as dual-modality guidance:  
\begin{equation}
    \label{eqn:similarity}
    \small
    \alpha^t = \mathrm{sim}( x_\mathrm{proto}^t, x_\mathrm{v}),\quad
    \beta^t = \mathrm{sim}(x_\mathrm{proto}^t, x_\mathrm{instruct}),
\end{equation}
where similarity is a measurement that matches current multimodal input and task-specific prompts and we exploit commonly used cosine similarity. With dual-modality guidance, the model has the ability to determine which prompts may boost the performance of evaluated task. We then select the prompts among $\{1,\cdots,t\}$ with $k$ largest similarity of multimodal supervision:
\begin{equation}
    \widetilde{X}_\mathrm{p}^t = X_\mathrm{p}^{\leq t} \circ \mathcal{I}_k\{\alpha^{\leq t} + \beta^{\leq t}\},
    \label{eqn:train-selection}
\end{equation}
where $\mathcal{I}_k:\mathbb{R}^{(M\times t)\times{d_t}}\rightarrow\mathbb{R}^{(M\times k)\times{d_t}}$ represents selecting the index with the largest $k$ elements, and $\circ$ means selecting according to index. Note that in order to optimize parameters of current task, prompts of current task are always selected during training and prompts belonging to the same task are always selected simultaneously.

In summary, we explicitly integrate prompts that are close to the feature distribution of current task into training procedure by utilizing supervision from both modalities that caters for LMMs, and therefore transfer previous knowledge to boost the performance of current task. 

\noindent \textbf{Training objectives.} 
During training, the inputs for continual learning of task $\mathcal{T}^t$ are prefixed with fused prompts $\widetilde{X}_\mathrm{p}^t$ described in Eqn.~\ref{eqn:train-selection}. As shown in Fig.~\ref{fig:detailed-structure}, parameters of large language model $\theta$ are frozen, and the introduced projection layer along with prompts corresponding to current task $\theta_\mathrm{p}^t=\{\theta_{X_\mathrm{P}^t},\theta_{\mathrm{Proj_p}}\}$ are trainable. The optimization target for task $\mathcal{T}^t$ is to find optimal parameters $\theta_\mathrm{p}^t$ that minimize the negative log-likelihood language loss for LMMs:
\begin{equation}
    \small
    \begin{aligned}
    &\quad \quad \mathcal{L}_{\mathrm{LMM}}^t(\theta_\mathrm{p}^t) = \mathbb{E}_{(X_\mathrm{v}^t, X_\mathrm{instruct}^t, y^t)\sim \mathcal{D}_t}\\
    &\Big{[}- \sum_{\ell=1}^L \log p(y^{\ell} | [\widetilde{X}_\mathrm{p}^t, X_\mathrm{v}, X_\mathrm{instruct},y^{<\ell}],\theta,\theta_p^1,\cdots, \theta_p^t)\Big{]}.
    \end{aligned}
\end{equation}
where $L$ is the length of each sample pair in the dataset. The projection layer is optimized to reserve prototype feature during training process. Since we are to maximum the dual-modality guidance to keep knowledge of current task, we additionally design a prototype similarity loss to optimize prototype features formulated as:
\begin{equation}
    \small
    \mathcal{L}_{\mathrm{Proto}}^t = \Big{[} 1- \mathrm{sim}(x_\mathrm{proto}^t, x_\mathrm{v}) \Big{]} +  \Big{[} 1- \mathrm{sim}(x_\mathrm{proto}^t, x_\mathrm{instruct}) \Big{]}.
\end{equation}

Total training objective is a sum of language loss and prototype similarity loss:
\begin{equation}
    \mathcal{L}_\mathrm{Total}^t = \mathcal{L}_{\mathrm{Proto}}^t + \mathcal{L}_{\mathrm{LMM}}^t.
\end{equation}
Parameters of current task are frozen afterwards to retain knowledge of learned tasks when learning new task.

\subsection{Inference: Dual-Modality Prompt Selection}
\label{sec:prompt-selection}
Another issue for MCIT is growing complexity with the number of tasks in evaluation. To handle this, we utilize dual-modality guidance for prompt selection in inference. As shown in middle of Fig.~\ref{fig:detailed-structure}, by selecting the most relevant prompts, we convert the problem of computational complexity from task numbers $\mathcal{O}(T)$ to selected prompt numbers $\mathcal{O}(k)$, which greatly reduces computation load and improves efficiency.

\vspace{0.2em}
\noindent \textbf{Prompt selection for complexity management.} After training on all sequential datasets, the crucial problem for evaluation is that the model has no ability to recognize which set of prompts promotes particular task and cannot manage computational complexity with simple ensembling leading to $\mathcal{O}(T)$ inference complexity. Intuitively, without access to data from older tasks, task-specific prompts should obtain cues for image-text distribution and be discriminant about which set of prompts counts during inference. To achieve this goal, we measure the similarity between image-text distribution of certain tasks and task-specific prompts employing dual-modality guidance. The representations of prototype features and multimodal feature are similar to Eqn.~\ref{eqn:proto} and Eqn.~\ref{eqn:similarity}, and differs in that selected prompts is determined among all $T$ set of prompts:
\begin{equation}
    \label{eqn:eval-selection}
    \widetilde{X}_\mathrm{p}^{\mathrm{eval}} = X_\mathrm{p}^{\leq T} \circ \mathcal{I}_k\{\alpha^{\leq T} + \beta^{\leq T}\}.
\end{equation}

\begin{table*}[ht]
    \centering
    \setlength{\tabcolsep}{3pt}
    \renewcommand\arraystretch{1.15}
    \resizebox{\linewidth}{!}{
    \begin{tabular}{c>{\raggedright\arraybackslash}p{2.5cm} >{\centering\arraybackslash}p{1.4cm} >{\centering\arraybackslash}p{1.4cm}>{\centering\arraybackslash}p{1.4cm} >{\centering\arraybackslash}p{1.4cm} >{\centering\arraybackslash}p{1.4cm} >{\centering\arraybackslash}p{1.4cm} >{\centering\arraybackslash}p{1.4cm} >{\centering\arraybackslash}p{1.4cm} >{\centering\arraybackslash}p{2cm}}
    \toprule[1.3pt]
        & {\hspace{1em}}Method & \small{\textbf{ScienceQA}} & \small{\textbf{TextVQA}} &\small{\textbf{ImageNet}}& \small{\textbf{GQA}} &\small{\textbf{VizWiz}}& \small{\textbf{REC}}& \small{\textbf{VQAV2}}& \small{\textbf{OCRVQA}} & \small{\textit{Average}} \\
        \midrule 
        &{\hspace{1em}}Multi-task& 81.43 & 61.36 &90.05& 60.67& 52.39& 66.14& 63.54& 61.28&67.10\\
        &{\hspace{1em}}Zero-shot & 67.92	&57.71&	47.37&	61.28	&45.17	&6.12&	52.03&	53.58&48.89\\
        \midrule
        \midrule
         \multirow{6}{*}{\rotatebox{90}{\textbf{Last}}} &{\hspace{1em}}Finetune& 26.00& 25.38& 28.51& 33.07 &26.52 &0.10 &40.00& 52.92 &29.06\\
        & {\hspace{1em}}\small{CODA-Prompt}&58.15&50.16&24.04&54.33&48.94&17.83&55.86&54.42&	45.46\\
        & {\hspace{1em}}Dualprompt&56.40&47.12&34.96&42.03&44.14&12.01&54.43&53.36&43.05\\
        & {\hspace{1em}}L2P&54.42&46.04&30.36&57.09&42.19&9.38&50.45&54.03&42.99 \\
        & {\hspace{1em}}MoELoRA&47.34& 32.91 &38.73 &37.15 &42.48& 0.97 &42.77& 57.50&37.48 \\ 
         &{\hspace{1em}}\textbf{Ours}&\bftab{68.42}&\bftab{56.40}&\bftab{41.13}&\bftab{61.11}&\bftab{50.13}&\bftab{36.69}&\bftab{66.90}&\bftab{59.68}&\bftab{55.06}~\textcolor{Maroon}{(\textbf{+9.60})}\\ 
         \midrule
        \multirow{6}{*}{\rotatebox{90}{\textbf{Avg}}}& {\hspace{1em}}Finetune& 13.79&15.74&9.08&28.84&15.20&0.06&40.00&-&17.53\\
        & {\hspace{1em}}\small{CODA-Prompt}&48.84&	47.17&18.74&50.77&42.68&15.43&55.86&-&39.93
\\		
        & {\hspace{1em}}Dualprompt&42.81&43.41&24.12&40.52&40.39&12.76&54.43&-&36.92\\
        & {\hspace{1em}}L2P &43.76&41.35&18.28&50.03&38.78&8.77&50.45&-&35.91\\
        &{\hspace{1em}}MoELoRA&39.12&27.10&20.01&40.65&28.72&1.36&42.77&-&28.53\\
         &{\hspace{1em}}\textbf{Ours}&\bftab{68.36}&\bftab{56.30}&\bftab{39.66}&\bftab{61.45}&\bftab{50.02}&\bftab{36.66}&\bftab{66.90}&-&\textbf{54.19}~\textcolor{Maroon}{(\textbf{+14.26})}\\
    \bottomrule[1.3pt]
    \end{tabular}}
    \vspace{-6pt}
    \caption{Comprehensive comparison of multimodal continual instruction tuning ability. Performance is measured with accuracy.}
    \vspace{-10pt}
    \label{tab:main-results}
\end{table*}

Intuitively, dual-modality guidance could serve as cues for selecting prompts that are helpful to current task in feature space, thereby converting computational complexity from $\mathcal{O}(T)$ to $\mathcal{O}(k)$. 

\vspace{0.2em}
\noindent \textbf{Response generation.} For each evaluated task, selected prompts $\widetilde{X}_\mathrm{p}^{\mathrm{eval}}$ with multimodal input $\{X_\mathrm{v}, X_\mathrm{instruct}\}$ are fed into LMM in a prefix way to generate answers:
\begin{equation}
    f([\widetilde{X}_\mathrm{p}^{\mathrm{eval}};X_\mathrm{v};X_\mathrm{instruct}]; \theta),
\end{equation}
where $[\cdot;\cdot]$ represents concatenation and $\widetilde{X}_\mathrm{p}^{\mathrm{eval}}$ is selected prompts through prompt selection. 

\vspace{0.2em}
\noindent \textbf{Remarks.} It can be concluded from above that the dual-modality guidance plays a crucial role in prompt learning for CIT and has two advantages: (1) help transfer knowledge from similar tasks to boost MCIT performance; (2) manage the inference speed as the time complexity is in proportion to the selected prompt numbers other than task numbers.

\section{Experiments}
\label{sec:experiment}
We apply LLaVA~\cite{liu2024visual} as base LMM, and CLIP-Large-336~\cite{radford2021learning} as vision and text encoder for dual-modality feature extraction. Prompts can be easily constructed by extending the vocabulary size of language tokenizer. Length for each prompt $M$ is set to 10. During prompt fusion and selection, we select 3 set of prompts. More implementation details can be found in Appendix~\ref{app:imple-detail}.

\vspace{0.2em}
\noindent \textbf{Datasets.} We employ CoIN~\cite{chen2024coin}, a MCIT benchmark with numerous vision-language instruction datasets to evaluate continual instruction tuning ability. It includes OCRVQA~\cite{mishra2019ocr}, GQA~\cite{hudson2019gqa}, 
ImageNet~\cite{deng2009imagenet}, ScienceQA~\cite{lu2022learn}, 
Vizwiz~\cite{gurari2018vizwiz}, 
TextVQA~\cite{singh2019towards},  VQAv2~\cite{goyal2017making} and RefCoco~\cite{mao2016generation,kazemzadeh2014referitgame}. Some of these datasets are visual question answering tasks of different fields, \eg, GQA for visual reasoning and ScienceQA for science knowledge, and others are classification~(ImageNet) and grounding~(RefCoco). Following CoIN, we perform continual instruction tuning in the order of ScienceQA, TextVQA, ImageNet, GQA, VizWiz, REC, VQAV2 and OCRVQA and evaluate the performance after each continual stage. 

\vspace{0.2em}
\noindent \textbf{Compared methods.} Apart from MoELoRA~\cite{dou2023loramoe}, we implement three advanced prompt-based continual learning methods including L2P~\cite{zhou2022learning}, Dualprompt~\cite{wang2022dualprompt} and CODA-Prompt~\cite{smith2023coda} in the architecture of LMM for comprehensive comparison. We try our best to get optimal results and briefly introduce compared method and hyper-parameters in Appendix~\ref{app:comparing-method}.

\vspace{0.2em}
\noindent \textbf{Evaluation metrics.} Denote that $A_{t,i}(i\leq t)$ is performance of task $i$ after training on task $t$~($T$ tasks in total). (1) For final performance evaluation, we measure each dataset using metrics \textbf{\textit{Last}}~(performance after sequential training on all tasks, \ie, $A_{T,i}, i=1,\cdots,T$) and \textbf{\textit{Avg}}~(average performance across MCIT procedure). (2) For time-dependent continuous evaluation, we evaluate continuous metrics at each incremental stage across all trained datasets. The metrics include \textbf{\textit{Backward Transfer~(B)}} and \textbf{\textit{Mean Accuracy~(M)}}. Zero-shot and multi-task are also reported to stand for the lower and upper bounds of the benchmark. Detailed explanations of these metrics are in Appendix~\ref{app:evaluation-metrics}.

\begin{table}[t]
    \centering
    \resizebox{\linewidth}{!}{
    \begin{tabular}{{c| >{\centering\arraybackslash}p{1.3cm} >{\centering\arraybackslash}p{1.3cm}>{\centering\arraybackslash}p{1.3cm}>{\centering\arraybackslash}p{1.3cm}}}
    \toprule[1.3pt]
        \diagbox{Metrics}{Guidance} & Image&Text&Dual&$\Delta$\\
    \midrule \midrule
        \textit{Last}  &51.95&50.39&55.06&\textcolor{Maroon}{+3.11}\\
         \textit{Avg} &50.35&49.02&54.19&\textcolor{Maroon}{+3.84}\\
    \bottomrule[1.3pt]
    \end{tabular}}
    \caption{Effectiveness of guidance from multimodal supervision. Dual-modality similarity guidance achieves the best results.}
    \vspace{-15pt}
    \label{tab:ablation-dual-guidance}
\end{table}

\subsection{Main Results}
\label{sec:results}
\noindent \textbf{Final continual performance.} Results of MCIT benchmark are shown in Tab.~\ref{tab:main-results}. It can be concluded that: \textbf{(1)} Existing LoRA-based and prompt-based methods shows limited promotion in MCIT, highlighting the necessity of specific methods for LMMs. By contrast, our method achieves remarkable improvements with substantial \textbf{+9.60\%} and \textbf{+14.26\%} gain, respectively. Notably, the results after sequential tuning even against multi-task training, strongly demonstrating the effectiveness of the dual-modality guided prompt learning framework. \textbf{(2)} When learning different types of tasks, our approach undergoes moderate performance drop and still gets competitive results other than losing the ability to respond to the task~(\eg, the performance of MoELoRA drops to near zero when evaluated on Grounding), indicating the continual learning ability of the proposed method. \textbf{(3)} \textit{Avg} of previous methods drop significantly compared with \textit{Last}, and our method has almost no degradation, consistently achieving superior performance across the continuous tuning. More comparison results can be found in Appendix~\ref{app:full-results}.

\begin{figure}[t]
    \centering
    \vspace{-8pt}
    \includegraphics[width=\linewidth]{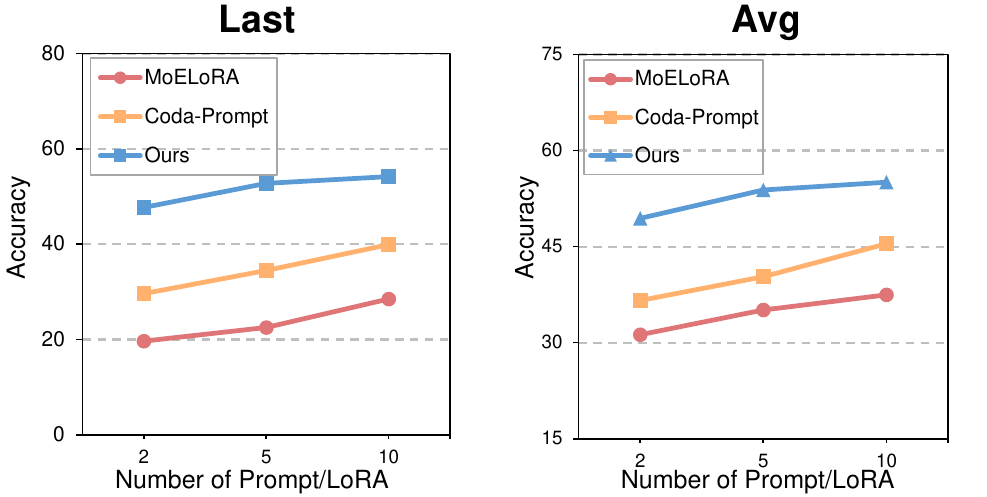}
    \vspace{-15pt}
    \caption{Impact of prompt/LoRA number. We implement different number of prompts for each task and different number of MoE for LoRA.}
    \label{fig:ablation-prompt-number}
    \vspace{-15pt}
\end{figure}

\begin{table*}[t]    
    \centering
    \setlength{\tabcolsep}{5pt}
    \renewcommand\arraystretch{1.3} 
    \resizebox{\linewidth}{!}{
    \begin{tabular}{l |cc cc cc cc cc cc cc |cc}
    \toprule[1.3pt]
        \multirow{2}{*}{Method} &\multicolumn{2}{c}{TextVQA} &\multicolumn{2}{c}{ImageNet}& \multicolumn{2}{c}{GQA} &\multicolumn{2}{c}{VizWiz}& \multicolumn{2}{c}{REC}&\multicolumn{2}{c}{VQAV2} & \multicolumn{2}{c}{OCRVQA} & \multicolumn{2}{c}{\textit{Average}}\\
        \cline{2-17}
        &$B_2 \downarrow$ & $M_2 \uparrow$ & $B_3 \downarrow$ & $M_3 \uparrow$ & $B_4 \downarrow$ & $M_4 \uparrow$ & $B_5 \downarrow$ & $M_5 \uparrow$ & $B_6 \downarrow$ & $M_6 \uparrow$ & $B_7 \downarrow$ & $M_7 \uparrow$ & $B_8 \downarrow$ & $M_8 \uparrow$ &  $B \downarrow$ &  $M \uparrow$\\ 
        \midrule
         Finetune & 44.30&44.14&65.53 &32.52&64.42&22.75&51.98&25.55&67.08&5.71&37.62&23.64&31.16&29.06&51.73&26.19\\
         CODA-Prompt& 11.54&57.88&27.70&34.05&14.38&43.44&12.78&42.76&12.72&39.28&14.05&39.42&7.27&45.46&14.34&43.18\\
         MoELoRA &41.31&43.13&52.47&34.08&32.76&41.71&33.81&37.71&41.41&25.59&30.80&34.34&26.12&37.48&36.95&36.29\\
         \textbf{Ours}& \bftab{6.55}&\bftab{64.50}&\bftab{4.40}&\bftab{56.34}&\bftab{3.16}&\bftab{57.63}&\bftab{4.51}&\bftab{54.15}&\bftab{3.98}&\bftab{50.96}&\bftab{2.02}&\bftab{54.07}&\bftab{1.41}&\bftab{55.06}&\textbf{3.72}~\textcolor[rgb]{0,0,0.81}{(\textbf{-10.6})}&\textbf{56.10}~\textcolor{Maroon}{(\textbf{+12.9})}\\ 
         
         \bottomrule[1.3pt]
    \end{tabular}}
    \vspace{-7pt}
    \caption{Continual performance metrics at each incremental stage. $B_t$ and $M_t$ stand for \textit{Backward Transfer} and  \textit{Mean Accuracy} at incremental stage $t$.}
    \label{tab:continuous-metrics}
    \vspace{-15pt}
\end{table*}

\begin{table}[t]
    \centering
    \setlength{\tabcolsep}{8pt}
    \renewcommand\arraystretch{1.15}
    \vspace{2pt}
    \resizebox{0.95\linewidth}{!}{
    \begin{tabular}{cc cc cc}
    \toprule[1.3pt]
        Fusion & Selection& Last& Avg& B&M\\
    \midrule \midrule
        &$\checkmark$&37.36&31.87&27.09&34.94\\
        $\checkmark$&&44.81&38.24&17.16&40.71\\
        \rowcolor[rgb] {1,1, 0.848} $\checkmark$ &$\checkmark$&55.06&54.19&3.72&56.10\\
    \bottomrule[1.3pt]
    \end{tabular}}
    \vspace{-4pt}
    \caption{Effectiveness of the proposed prompt selection and fusion for continual learning. Without prompt selection, we concatenate all prompts like Progressive Prompts~\cite{razdaibiedina2023progressive}.}
    \label{tab:ablation-selection-fusion}
    \vspace{-15pt}
\end{table}

\noindent \textbf{Continuous continual performance.} We also evaluate continuous metrics at each incremental stage in Tab.~\ref{tab:continuous-metrics} to examine time-variant multimodal continual instruction tuning performance. Concretely, compared with previous methods, our method is especially effective in alleviating catastrophic forgetting~(BWT) to the most~(\textbf{10.6\%} mitigation) and also gets continuous promotion in across MCIT~(\textbf{12.9\%}). It is evident that our method outperforms state-of-the-art prompt-based method CODA-Prompt and LoRA-base method MoELoRA by a substantial margin with respect to both anti-forgetting and enhancing mean accuracy.

\subsection{Ablation Study}
We conduct numerous ablation studies to carefully validate the effectiveness of components and hyper-parameters in the proposed method.

\noindent \textbf{Effectiveness of dual-modality guidance.} We develop the unique dual-modality guidance tailored for LMMs with multimodal information. To demonstrate the importance of multimodal guidance, we replace it with single-modality guidance. 
It is evident in Tab.~\ref{tab:ablation-dual-guidance} that either image or text information solely performs inferior to the proposed multimodal strategy, and vision information from multimodal dataset plays an inescapable function in guiding MCIT especially in datasets that rely heavily on image scenes. This strongly showcases that our method improves performance of MCIT by retaining robust and reliable prototype features in multimodal feature space and therefore contributes to all continuous tasks.

\noindent \textbf{Prompt fusion and selection.}
We design the dual-modality guidance for knowledge transfer and complexity management, respectively. To validate the effectiveness of each proposed mechanism, we ablate each of them to demonstrate their usefulness. It is shown in Tab.~\ref{tab:ablation-selection-fusion} that both of them play a key role in the framework and lacking either of them causes severe performance drop. Specifically, multi-task prompt fusion is significant in promoting continual learning in the form of knowledge transfer. Besides, without selection, knowledge of different types of tasks would confuse the model and lead to performance drop. All results strongly highlight the effectiveness of dual-modality guidance in MCIT framework.

\noindent \textbf{Number of prompts.} The number of prompts represents prototype features in aligned image-text space. We vary both numbers of prompt and LoRA in different methods to investigate the influence of prompt numbers. Results in Fig.~\ref{fig:ablation-prompt-number} elucidate that increasing prompt numbers brings slight performance improvement. Considering both effectiveness and efficiency, we set the number of prompts for each task to 10 and do not further expand the quantity.

\begin{figure}[t]
    \centering
    \hspace{-15pt}
    \includegraphics[width=1.05\linewidth]{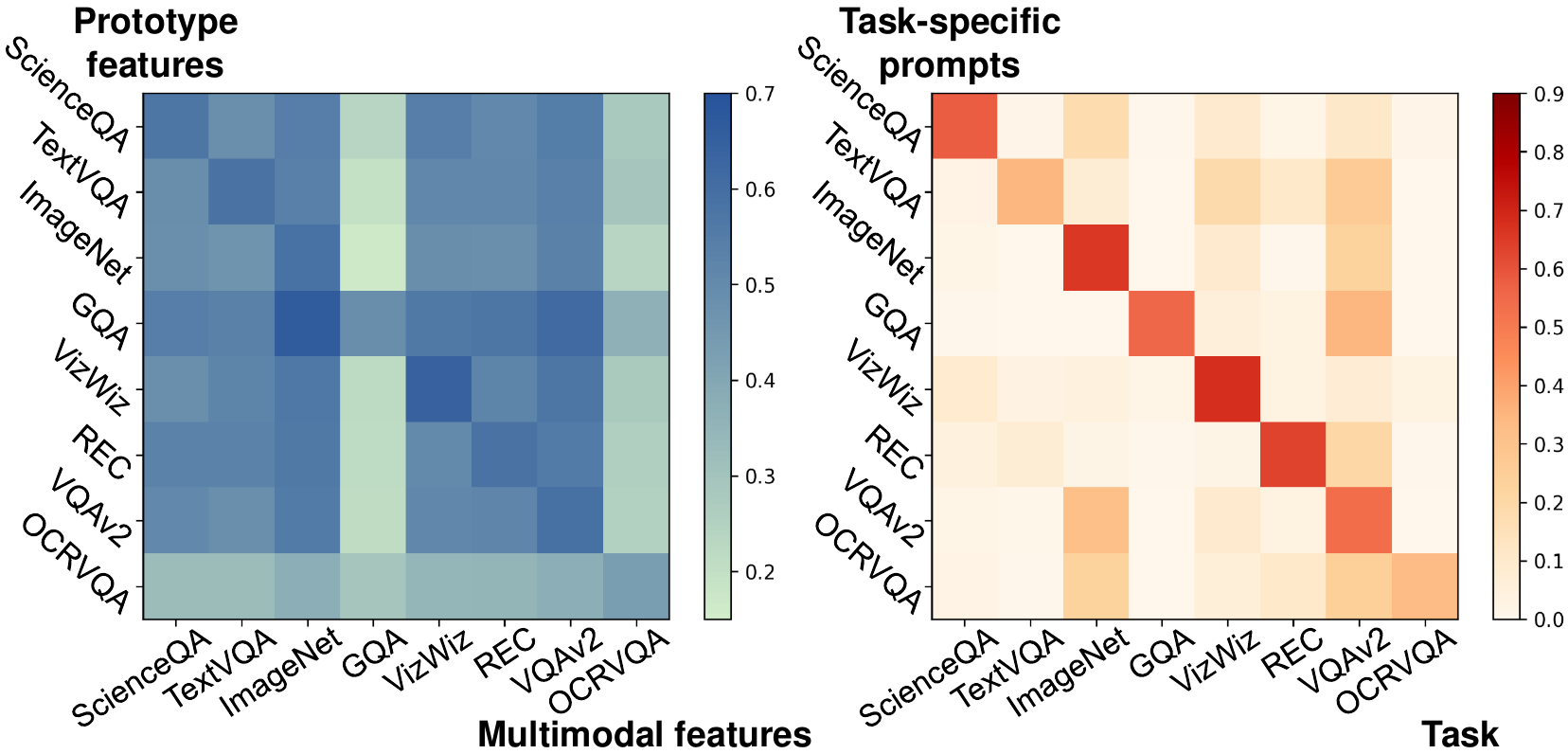}
    \vspace{-15pt}
    \caption{\textit{Left:} Similarity between prototype features and multimodal task features. Larger value indicates more similar distribution. \textit{Right:} Selection probability of each task from prompts. Results are percentage so the sum of rows equals one. Zoom in for better view.}
    \label{fig:ablation-heatmap}
    \vspace{-13pt}
\end{figure}

\begin{figure*}[t]
    \centering
    \includegraphics[width=0.97\linewidth]{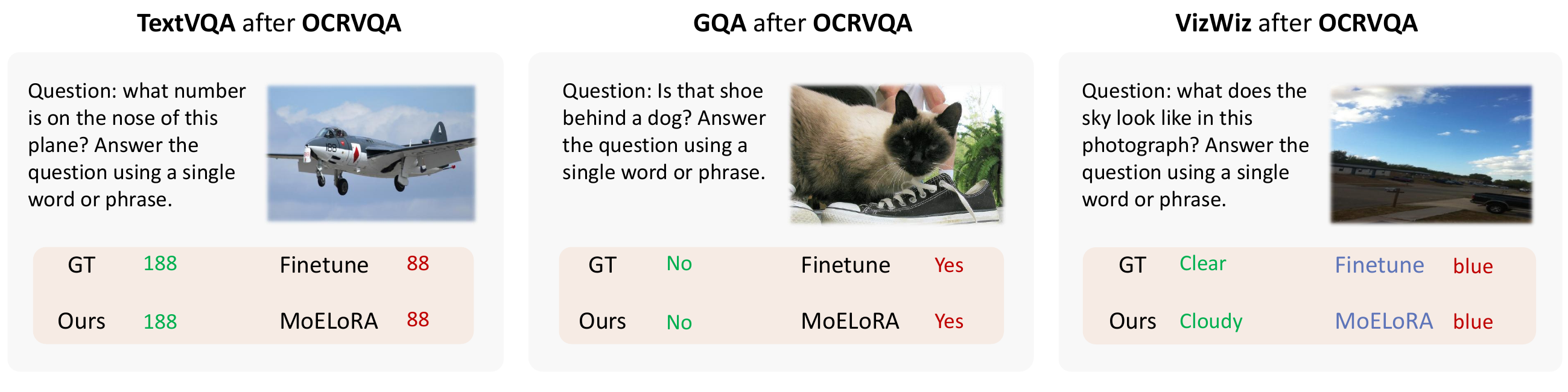}
    \vspace{-10pt}
    \caption{Multimodal continual instruction tuning responses of several examples from TextVQA, GQA and VizWiz after fine-tuning on OCRVQA. Our method successfully mitigate forgetting and gives correct answers.}
    \vspace{-12pt}
    \label{fig:visualization}
\end{figure*}

\subsection{Further Analysis}
\label{sec:analysis}
\textbf{Efficiency comparison.} As prompt learning serves as another way to efficiently fine-tune large models, we compare our approach with LoRA-based~\cite{chen2024coin} and prompt-based~\cite{smith2023coda} method in terms of additional parameters, inference latency and GPU memory consumption to assess the efficiency. Tab.~\ref{tab:efficiency-comparison} reveals that our strategy achieves better results with lower memory, inference latency and trainable parameters. Specifically, we merely train \textbf{0.27\%} of total parameters, which is \textbf{5\%} of MoELoRA and \textbf{13\%} of CODA-Prompt. Therefore, compared with baseline, our method achieves faster inference speed~($\times$\textbf{1.42}), reduces training time~($\times$\textbf{0.35}) and GPU memory consumption~($\times$\textbf{0.92}), firmly substantiating the efficiency of our approach. The achievements can be attributed to simple prompt learning implementation and the prompt selection module that manages the computational complexity, consequently improving the inference efficiency.

\begin{table}[t]
    \centering
    \setlength{\tabcolsep}{6pt}
    \renewcommand\arraystretch{1.25}
    \vspace{5pt}
    \resizebox{\linewidth}{!}{
    \begin{tabular}{ccccc}
    \toprule[1.3pt]
        \multirow{2}{*}{Method} & Memory & Training  & Trainable &Throughput\\
        &($M$) & (Hour)&Param&(Token/$s$)\\
    \midrule\midrule
        MoELoRA &16784&10.74 & 4.73\% &2.41  \\
        CODA-Prompt & 16073 &5.12&  1.97\%&2.90\\
        \bftab{Ours}&\bftab{15517} &\bftab{3.81} &\bftab{0.27\%}&\bftab{3.43} \\
        \midrule
            $\Delta$ & \textbf{\textcolor[rgb]{0,0,0.81}{$\times$0.92}} & \textbf{\textcolor[rgb]{0,0,0.81}{$\times$0.35}} & \textbf{\textcolor[rgb]{0,0,0.81}{$\times$0.05}}&\textbf{\textcolor{Maroon}{$\times$1.42}}\\
    \bottomrule[1.3pt]
    \end{tabular}}
    \vspace{-5pt}
    \caption{Efficiency comparison of typical LoRA and prompt-based method with respect to GPU consumption, speed and trainable parameters. We average the training time for one epoch across datasets.}
    \vspace{-13pt}
    \label{tab:efficiency-comparison}
\end{table}

\noindent \textbf{Similarity of dual-modality features.} 
The ability of our framework to learn continually is largely guaranteed by the prompt selection module and prototype features represented in vision-language feature space. To further analyze the effectiveness of the dual-modality guidance tailored for LMMs, we calculate the similarity matrix between prototype features and multimodal task features. In Fig.~\ref{fig:ablation-heatmap}, the similarity heatmap vividly illustrates the vision-language distributions of continual learning tasks. First, multimodal features of a few tasks are similar, indicating that most multimodal tasks share common sense and can promote each other mutually. However, some tasks, such as GQA and OCRVQA, are not similar to other tasks, which may be due to their task-specific ability not needed by other common tasks~(visual reasoning for GQA and OCR for OCRVQA); second, the similarity is asymmetric, which may be attributed to their task inclusion relationship. For instance, GQA requires higher-level reasoning ability, while some other tasks may merely need to answer questions based on visual-language information. Therefore, features of GQA task~(more basically) are similar to other tasks, but other tasks~(more specifically) are not similar to the prototype of GQA. The visualization of dual-modality features exhibits the connection between prior obtained knowledge~(prototype features) and given task~(multimodal task features), and therefore contributes fundamentally to continual learning ability of LMMs.

\noindent \textbf{Selection of prompts.} To have an intuitive understanding of prompt selection module in addition to soft distribution construction, we report selection results of each previous task in percentage under MCIT setting to figure out the actual selection of prompts during inference. The results in right of Fig.~\ref{fig:ablation-heatmap} expose that the proposed module correctly matches and prioritizes prompts of the corresponding task as prefixes to enhance MCIT. Moreover, the module also selects prompts from similar type of tasks, which also enhances performance. This strongly indicates that knowledge transfer in tasks of the same type can mutually promote the performance, and our method leverages this characteristic excellently, demonstrating the robustness and usefulness of the learned prototype features.

\noindent \textbf{Visualization.} Fig.~\ref{fig:visualization} provides examples during MCIT procedure to explicitly illustrate the effectiveness of our method. Specifically, existing methods fail on challenging multimodal generative tasks especially dependent on vision information. By contrast, our method can maintain performance on diverse previous tasks, keep knowledge from multimodality and answer challenging questions requiring comprehensive understanding correctly. For example, in TextVQA, the model identifies the specific part location of objects~(nose of the plane) and overcomes occlusion; in GQA, it successfully distinguishes spatial orientation and therefore identifies objects. Moreover, it also deduces appropriate answers with analogous meanings to the ground truth based on image and text questions~(\eg, cloudy and clear in VizWiz). The visualizations demonstrate that based on the retained multimodal knowledge, our model gives the correct answer for diverse generative tasks, outperforming traditional existing continual instruction learning methods without design for vision information.

\section{Conclusion}
In this paper, we overcome the obstacle of continual learning tailored for LMMs with efficiency, and propose to exploit efficient prompt learning for continually learning image-text generative tasks while retaining knowledge of older tasks from multimodal supervision. Specifically, we construct a set of prompts for each task to represent task-specific knowledge in feature space. Building upon dual-modality guidance, we propose prompt fusion to enhance the performance from knowledge transfer and leverage prompt selection to manage the computational complexity of the model. Comprehensive experiments and analyses validate the effectiveness and efficiency of our framework.

\section*{Limitations}
In this article, we propose ModalPrompt, an approach that exploits effective prompt fusion and selection with dual-modality guidance to retain performance in multimodal continual instruction tuning. While obtaining impressive continual learning performance, our method only retains knowledge of learned tasks and falls short of enhancing unseen tasks. However, we argue that it is an underexplored field as multimodal continual learning itself is not fully investigated. We will treat promoting forward transfer as future direction. Also, we believe that our model is generalizable and versatile, and plan to scale model size and application to other LMMs in future works.

\section*{Ethical Impact}
We are committed to safeguarding intellectual property rights and complying with all applicable laws and regulations. The multimodal instructions included in our experiments are open-sourced from publicly available materials. We are dedicated to research purpose and are not intended for any commercial use.

\section*{Acknowledgments}
\begin{sloppypar}
This work was supported by the National Natural Science Foundation of China~(62222609), National Science and Technology Major Project~(2022ZD0116500), CAS Project for Young Scientists in Basic Research~(YSBR-083), and the InnoHK program.
\end{sloppypar}

\bibliography{custom}

\cleardoublepage
\appendix

\section{Details of Evaluation Metrics}
\label{app:evaluation-metrics}
We give a thorough definition and explanation of the evaluation metrics used in the main experiments.

\noindent (1) Average: In addition to \textit{Last}, which focuses on performance after tuning on all datasets, we propose to average performance throughout the entire tuning process. $\mathrm{Avg}_i = \frac{1}{T-i}\sum_{t=i+1}^{T}A_{t,i}, i=1,2,\dots,T-1.$ It measures the absolute performance of each data across the sequential tuning. It is vital to keep the performance from dropping severely when the fine-tuning task varies greatly. 

\noindent (2) Backward Transfer~(BWT): It reflects the relative variation between current performance and direct tuning performance, measuring the catastrophic forgetting on all tasks. $B_t =\frac{1}{t-1}\sum_{i=1}^{t-1}(A_{i,i} - A_{t,i}), t=2,\cdots,T$. Lower BWT represents better anti-catastrophic forgetting performance.
 
\noindent (3) Mean Accuracy~(MA): $M_t = \frac{1}{t}\sum_{i=1}^tA_{t,i}$. It measures the average performance of all tasks at each incremental stage and is introduced to evaluate continual learning ability of all previous tasks. Higher MA stands for better continual learning ability. The above two metrics are averaged across all data on each incremental stage except the first one, \ie, $t=2,\dots,T$.

\section{More Details of Experimental Settings}
\label{app:imple-detail}
\noindent \textbf{Continual instruction templates.} For continual instruction tuning, the instructions for each datasets is shown in Tab.~\ref{tab:instruction-template}. Large language model concatenates instructions with image-text pairs in datasets to generate response accordingly.

\vspace{0.2em}
\noindent \textbf{Additional Implementation Details.} Our framework is constructed depending on deepspeed repository~\footnote{\href{https://github.com/microsoft/DeepSpeed}{https://github.com/microsoft/DeepSpeed}} and Visual Instruction Tuning~\footnote{\href{https://github.com/haotian-liu/LLaVA}{https://github.com/haotian-liu/LLaVA}}. The instructions are from the repository of CoIN~\footnote{\href{https://github.com/zackschen/CoIN}{https://github.com/zackschen/CoIN}}. In evaluation of ImageNet, we give option choices for each question-answer pairs to avoid inaccurate descriptions. During training, all experiments are conducted on 48G NVIDIA A6000 and batch size is adaptively adjusted to maximize the memory utilization.

\begin{table}[t]
    \centering
    \setlength{\tabcolsep}{6pt}
    \renewcommand\arraystretch{1.15}
    \resizebox{\linewidth}{!}{
    \begin{tabular}{c|c}
    \toprule[1.3pt]
        \bftab{Dataset} & \bftab{Instruction} \\\midrule
         ScienceQA  &\makecell[c]{Answer with the option's letter from \\ the given choices directly.}   \\\midrule
         TextVQA &  \makecell[c]{Answer the question using a single \\ word or phrase.}\\\midrule
         ImageNet&  \makecell[c]{What is the object in the image? \\ Answer the question using a single word \\ or phrase.}  \\\midrule
         GQA &\makecell[c]{Answer the question using a single \\ word or phrase.} \\ \midrule
         VizWiz& \makecell[c]{Answer the question using a single \\ word or phrase.}  \\\midrule
         Grounding& \makecell[c]{Please provide the bounding box coordinate \\ of the region this sentence \\ describes: \texttt{<description>}.}\\\midrule
         VQAv2&\makecell[c]{Answer the question using a single \\ word or phrase.}\\\midrule
         OCR-VQA& \makecell[c]{Answer the question using a single \\ word or phrase.} \\
         \bottomrule[1.3pt]
    \end{tabular}}
    \caption{Instructions for each evaluated dataset.}
    \label{tab:instruction-template}
    \vspace{-12pt}
\end{table}

\begin{table*}[t]
    \centering
    \resizebox{0.95\linewidth}{!}{
    \begin{tabular}{ccc|cc cccc|c}
        \toprule[1.3pt]
        &Vision Encoder&Text Encoder&ScienceQA&TextVQA& GQA &VizWiz& VQAv2& OCRVQA&Average\\
        \midrule
        \multirow{3}{*}{\rotatebox{90}{\textbf{Last}}}&CLIP/L&CLIP/L& 67.82&56.41&60.76&51.08&66.93&59.52&60.42\\
        &CLIP/L&BPE& 62.42&53.73&62.27&45.92&65.11&62.49&58.66\\
        &CLIP/G&CLIP/G&68.40&55.94&62.16&50.16&68.54&60.62&60.97\\
        \midrule
        \multirow{3}{*}{\rotatebox{90}{\textbf{Avg}}} &CLIP/L&CLIP/L&67.41 &56.37&59.57&50.29&66.93&-&60.11\\
        &CLIP/L&BPE&        61.18&54.04&61.56&43.92&65.11&-&57.16 \\
        &CLIP/G&CLIP/G&68.29&56.11&62.28&49.97&68.54&-&61.04\\
        \bottomrule[1.3pt]
    \end{tabular}}
    \vspace{-5pt}
    \caption{Influence of different encoders for continual instruction tuning on a subset.}
    \vspace{-15pt}
    \label{tab:encoder}
\end{table*}

\section{More Experimental Results}
\label{app:full-results}
\textbf{Full continual instruction tuning results.} We showcase brief results in the main results. We provide detailed continual instruction tuning performance during evaluation at each incremental stage. Upper, middle and bottom of Tab.~\ref{tab:compare-full-results-1} are full result comparison of different LMM continual instruction tuning approaches, including Finetune, MoELoRA and Ours. It can be concluded that our method achieves consistent and significant improvements against previous LoRA based method, validating the effectiveness of our method. Additional results of prompt-based methods are also shown in Tab.~\ref{tab:compare-full-results-2}. It can be concluded that compared with prompt-based methods, our method also obtains substantial promotion, further certificating the utility of our approach.

\noindent \textbf{Influence of different encoders.} As stated in the paper, the dual-modality guidance coming from frozen CLIP-L plays a vital role in the proposed approach as the multimodal distribution from input is crucial to prompt selection and knowledge transfer. To explore the influence of different encoders, we replace text encoder with simple BPE tokenizer, and employ stronger vision encoder CLIP/G, respectively. Conclusion from Tab.~\ref{tab:encoder} is that stronger encoder is slightly better. Yet, we conclude that \textbf{(1)} guidance merely selects but not represents knowledge and CLIP/L is fairly effective, which can be validated by Fig.~\ref{fig:ablation-prompt-number} and Tab.~\ref{tab:encoder}; \textbf{(2)} as CLIP/L and image features have already been used in LLaVA, cost of computing and storage of CLIP-L is extremely small. Therefore, we use CLIP-L in current structure.

\vspace{0.2em}
\noindent \textbf{Comparison with more methods.} We compare our method with broader continual learning approaches including Model Tailor~\cite{zhu2024model}, LWF~\cite{li2017learning} and EWC~\cite{kirkpatrick2017overcoming}. Model Tailor maintains continual learning ability by calculating and enhancing important parameters for downstream tasks. It mainly focuses on reducing forgetting when fine-tuning a small number of downstream tasks and is not specifically designed for sequential adaptation across various tasks. As shown in Tab.~\ref{tab:add-results}, when encountering larger continual instruction tuning benchmarks, our method exhibits robust and better performance on the long continual learning process. Also, when compared with regularization-based approaches like LWF and EWC, our method obtains substantial improvements. These additional comparison comprehensively shows the advantage and effectiveness of our proposed method.

\begin{table}[H]
    \centering
    \renewcommand\arraystretch{1.25}
    \resizebox{\linewidth}{!}{
    \begin{tabular}{lccccccccc}
    \toprule[1.3pt]
    Method&\texttt{\textbf{S}}&\texttt{\textbf{T}}& \texttt{\textbf{I}}&\texttt{\textbf{G}}&\texttt{\textbf{V}}&\texttt{\textbf{R}}&\texttt{\textbf{Q}}&\texttt{\textbf{O}}&\texttt{\textit{Avg}}\\
    \hline
    LWF&57.42&53.01&31.03&47.10&40.06&17.08&52.17&53.44&43.91\\
    EWC&59.04&52.21&31.06&51.86&42.34&14.36&53.04&53.86&44.72\\
    Model Tailor&\bftab{77.01}&44.09&26.33&47.28&37.16&25.40&54.06&56.73&46.01\\
    Ours&68.42&\bftab{56.40}&\bftab{41.13}&\bftab{61.11}&\bftab{50.13}&\bftab{36.69}&\bftab{66.90}&\bftab{59.68}&\bftab{55.06}\\
        \bottomrule[1.3pt]
    \end{tabular}}
    \caption{Additional comparison results with broader methods. \texttt{\textbf{S}:}~\texttt{\textbf{S}cienceQA}, \texttt{\textbf{T}:}~\texttt{\textbf{T}extVQA}, \texttt{\textbf{I}:}~\texttt{\textbf{I}mageNet}, \texttt{\textbf{G}:}~\texttt{\textbf{G}QA}, \texttt{\textbf{V}:}~\texttt{\textbf{V}izWiz}, \texttt{\textbf{R}:}~\texttt{\textbf{R}EC},
    \texttt{\textbf{Q}:}~\texttt{V\textbf{Q}AV2},
    \texttt{\textbf{O}:}~\texttt{\textbf{O}CRVQA}.}
    \label{tab:add-results}
\end{table}

\begin{figure}[t]
    \centering
    \includegraphics[width=0.9\linewidth]{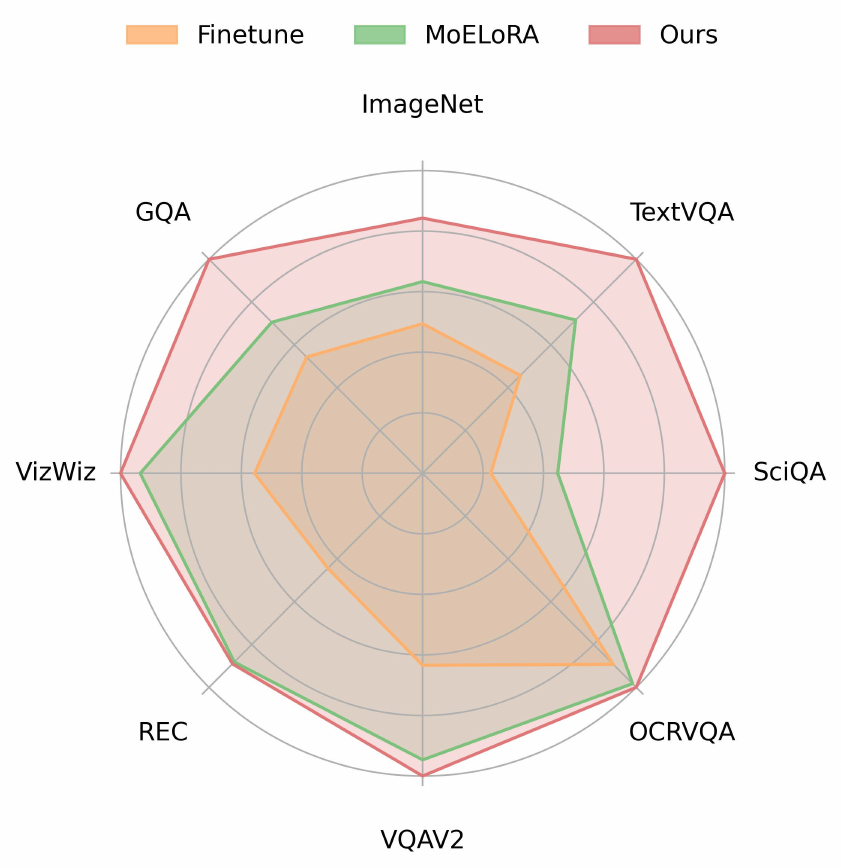}
    \caption{Results on Qwen-VL architecture.}
    \label{fig:qwen}
    \vspace{-10pt}
\end{figure}

\vspace{0.2em}
\noindent \textbf{Experiment on other LMMs.} We additionally conduct experiments on Qwen-VL~\cite{bai2023qwenvl}, which integrates Q-former to align vision features with language models, to validate the effectiveness of the method on different LMM architectures. Consistent and substantial improvements in Fig.~\ref{fig:qwen} certificate the adaptability and scalability of our method on different architectures, strongly demonstrating the extensibility and generalizability of the proposed method across different LMM architectures.

\section{Comparing Methods}
\label{app:comparing-method}
We briefly introduce methodology of comparing approaches and then show the hyperparameters of each method. For practical implementation of existing methods, we have tried our best to get optimal results under fair comparison. Hyperparameters not mentioned are set by default. 

\noindent \textbf{MoELoRA} leverages experts and gate function to activate part of parameters for each input. It learns knowledge of different tasks during training and mitigates forgetting in evaluation. We use 8 mixture of experts with rank $r=16$. 

\noindent \textbf{L2P} generates a pool of prompts in memory space. It manages task-invariant and task-specific knowledge in an explicit way of selecting relevant prompts for evaluation. We use pool size $M=10$ with each length of each prompt $L_p=10$ and select $N=3$ for each task.

\noindent \textbf{Dualprompt} employs general prompt and expert prompt to encode task-invariant and task-specific instructions, and attach them to different layers of transformer block to meet the demand of knowledge restoration. We select number of general prompts $L_g=3$ and number of expert prompts $L_e=10$ practically.

\noindent \textbf{CODA-Prompt} proposes to learn a
set of input-conditioned prompts for rehearsal-free continual learning. We use pool size $M=10$ and length of each prompt $L_p=10$.

\begin{table*}[ht]
    \centering
    \renewcommand\arraystretch{1.25} 
    \resizebox{\linewidth}{!}{
    \begin{tabular}{cccc ccccc}
    \toprule[1.3pt]
        \bftab{Finetune} & ScienceQA &TextVQA &ImageNet& GQA &VizWiz& REC& VQAV2& OCRVQA  \\
          ScienceQA  & 82.45  \\
          TextVQA  & 38.15&50.14 \\
          ImageNet & 0.96&0.58&96.03\\
          GQA &   13.91&15.78&5.67&55.65\\
          VizWiz  & 8.46&25.17&4.60&38.12&51.42\\
          REC     & 0.00&0.00&0.00&0.27&0.00&34.00\\
          VQAV2   &9.10&27.58&6.62&43.92&19.10&0.03&59.17\\
          OCRVQA &26.00&25.38&28.51&33.07&26.52&0.10&40.00&52.92\\
    \midrule
    \midrule
    \bftab{MoELoRA} & ScienceQA &TextVQA &ImageNet& GQA &VizWiz& REC& VQAV2& OCRVQA  \\
          ScienceQA  & 75.78  \\
          TextVQA  & 34.47 & 51.80 \\
          ImageNet & 22.61& 0.04&  79.60\\
          GQA &   32.37 &34.04 & 42.48 & 57.95 \\
          VizWiz  & 45.32&38.13 & 2.63 & 43.80&58.70 \\
          REC     & 58.76&9.08& 5.64 & 31.87&11.45&36.77\\
          VQAV2   &33.01& 48.42& 10.61 &49.78&32.23&1.75 &64.58 \\
          OCRVQA &47.34 &32.91 & 38.73 & 37.15 &42.48&0.97 &42.77& 57.50\\
    \midrule
    \midrule
        \bftab{ModalPrompt} & ScienceQA &TextVQA &ImageNet& GQA &VizWiz& REC& VQAV2& OCRVQA  \\
          ScienceQA  & 77.05  \\
          TextVQA  & 70.50 & 58.50 \\
          ImageNet & 68.57 & 58.18 & 42.26 \\
          GQA &   68.82 & 56.08 & 43.43 &  62.17 \\
          VizWiz  & 67.48 & 55.05 & 37.60 & 61.81 & 48.81 \\
          REC     & 66.58 & 55.68 & 35.92 & 61.95 & 48.74 &  36.88 \\
          VQAV2   &68.12&56.43& 40.22 & 60.92 & 51.19 & 36.63 & 64.99 \\
          OCRVQA &68.42&56.40& 41.13 & 61.11 & 50.13 & 36.69 & 66.90 & 59.68\\
    \bottomrule[1.3pt]
    \end{tabular}}
    \caption{Detail continual instruction tuning results of Finetune, MoELoRA and our method.}
    \label{tab:compare-full-results-1}
\end{table*}

\begin{table*}[ht]
    \centering
    \renewcommand\arraystretch{1.25} 
    \resizebox{\linewidth}{!}{
    \begin{tabular}{cccc ccccc}
    \toprule[1.3pt]
        \bftab{L2P} & ScienceQA &TextVQA &ImageNet& GQA &VizWiz& REC& VQAV2& OCRVQA  \\
          ScienceQA  & 72.83  \\
          TextVQA  & 68.07&57.16		\\
          ImageNet & 32.05&26.73&39.43	\\
          GQA &   47.53&46.02&18.03&60.47	\\
          VizWiz  & 65.94&37.68&1.72&56.29&47.90\\
          REC     & 5.74&42.96&33.92&39.44&39.64&1.87		\\
          VQAV2   &32.57&48.65&7.41&47.32&34.52&8.17&59.40	\\
          OCRVQA &54.42&46.04&30.36&57.09&42.19&9.38&50.45&54.03\\
    \midrule
    \midrule
    \bftab{Dualprompt}& ScienceQA &TextVQA &ImageNet& GQA &VizWiz& REC& VQAV2& OCRVQA  \\
          ScienceQA  & 67.16	\\					
          TextVQA  & 52.20&53.12\\
          ImageNet & 28.49&24.77&46.40\\
          GQA &   49.70&47.94&12.06&55.10\\
          VizWiz  & 57.88&51.17&21.34&48.03&51.62\\
          REC     & 18.27&39.64&29.77&44.06&35.97&30.82\\
          VQAV2   &36.77&49.85&22.48&27.96&41.08&13.51&61.27	\\
          OCRVQA &56.40&47.12&34.96&42.03&44.14&12.01&54.43&53.36\\
    \midrule
    \midrule
    \bftab{CODA-Prompt} & ScienceQA &TextVQA &ImageNet& GQA &VizWiz& REC& VQAV2& OCRVQA  \\
          ScienceQA  & 70.26	\\									
          TextVQA  & 58.72&57.05 \\
          ImageNet & 36.96&34.95&30.26	\\
          GQA &  50.78&53.52&10.12&59.35	\\
          VizWiz  & 55.37&47.21&6.78&56.43&48.01	 \\
          REC     & 33.56&47.67&32.07&44.62&43.39&34.42		\\
          VQAV2   &48.34&49.54&20.72&47.72&35.73&13.03&60.87	 \\
          OCRVQA &58.15&50.16&24.04&54.33&48.94&17.83&55.86&54.42\\
    \bottomrule[1.3pt]
    \end{tabular}}
    \caption{Detail continual instruction tuning results of prompt-based methods, inluding L2P, Dualprompt and CODA-Prompt.}
    \label{tab:compare-full-results-2}
\end{table*}

\end{document}